# FC Portugal 3D Simulation Team: Team Description Paper 2020

Nuno Lau[1,3], Luís Paulo Reis[2,4], David Simões[1,3], Mohammadreza Kasaei[3],
Miguel Abreu[2,4], Tiago Silva[4], Francisco Resende[1,3],

nunolau@ua.pt, lpreis@fe.up.pt, david.simoes@ua.pt, s.mohammadreza.kasaei@gmail.com,
{m.abreu,up201402841}@fe.up.pt, francisco.resende@ua.pt
[1]DETI/UA – Electronics, Telecommunications and Informatics Dep., University of Aveiro, Portugal
[2]DEI/FEUP – Informatics Engineering Dep., Faculty of Engineering of the University of Porto, Portugal
[3]IEETA – Institute of Electronics and Telematics Engineering of Aveiro, Portugal
[4]LIACC – Artificial Intelligence and Computer Science Lab., University of Porto, Portugal
http://www.ieeta.pt/robocup/

**Abstract.** The FC Portugal 3D team is developed upon the structure of our previous Simulation league 2D/3D teams and our standard platform league team. Our research concerning the robot low-level skills is focused on developing behaviors that may be applied on real robots with minimal adaptation using model-based approaches. Our research on high-level soccer coordination methodologies and team playing is mainly focused on the adaptation of previously developed methodologies from our 2D soccer teams to the 3D humanoid environment and on creating new coordination methodologies based on the previously developed ones. The research-oriented development of our team has been pushing it to be one of the most competitive over the years (World champion in 2000 and Coach Champion in 2002, European champion in 2000 and 2001, Coach 2nd place in 2003 and 2004, European champion in Rescue Simulation and Simulation 3D in 2006, World Champion in Simulation 3D in Bremen 2006 and European champion in 2007, 2012, 2013, 2014 and 2015). This paper describes some of the main innovations of our 3D simulation league team during the last years. A new generic framework for reinforcement learning tasks has also been developed. This paper also includes general information related to the design of our agent architecture. The current research is focused on improving the above-mentioned framework by developing new learning algorithms to optimize low-level skills, such as running and sprinting. We are also trying to increase student contact by providing reinforcement learning assignments to be completed using our new framework, which exposes a simple interface without sharing low-level implementation details.

## 1. Introduction

FC Portugal was built upon the low-level skills research conducted during previous years. Although there is still space for improvement in FC Portugal low-level skills, we feel that we currently have a very performing set of these skills. Our research on developing low-level behaviors is mainly focused on approaches which can also be applied on the real robot with minimal adaptation. In this matter we developed low-level skills using model-based approaches, in which the stability of humanoid behavior is modeled using physical systems. Control the stability dynamics of a humanoid robot is still challenging and it is one of the main research directions of our team. In Section 4, we will explain our approaches to develop robust and agile soccer low-level skills.

As another main research direction, we are also focused on the high-level decision and cooperation mechanisms of our agents. For RoboCup 3D soccer simulation competition that was based on spheres (from 2004 to 2006), the decisive factor (like in the 2D competition) was the high-level reasoning capacities of the players and not their low-level skills. Thus we worked mainly on high-level coordination methodologies for our previous teams. Since 2007 humanoid agents have been introduced in the 3D Simulation league, but the number of agents has been kept small until 2011. During this period research in coordination was not very important in the 3D league. Developing efficient low-level skills, contrarily to what should be the research focus of the simulation league, has been the main decisive factor in the 3D league, during this period. However, in 2011 the number of agents has increased to 9, and in 2012 teams were composed by 11 players making finally coordination, a very important issue for the efficiency of the team.

Our research on high-level soccer coordination methodologies and team playing is mainly focused on the adaptation of previously developed methodologies from our 2D soccer teams [1, 2, 3, 4, 5] to the 3D humanoid environment and on creating new coordination methodologies based on the previously developed ones. In our 2D teams, which participated in RoboCup since 2000 with very good results, we have introduced several concepts and algorithms covering a broad spectrum of the soccer simulation research challenges. From coordination techniques such as Tactics, Formations, Dynamic Positioning and Role Exchange, Situation Based Strategic Positioning and Intelligent Perception to Optimization based low-level skills, Visual Debugging and Coaching, the number of research aspects FC Portugal has been working on is quite extensive [1, 2, 3, 4, 5].

Several interesting topics were opened by the introduction of humanoid agents, including in the use of learning and optimization techniques for developing efficient both high-level and low-level skills. In previous work, we have introduced methods for developing very efficient low-level skills using optimization techniques [1, 6]. Recently, we have developed a new learning framework in which several optimization techniques have been included such as hill climbing (HC), tabu search (TS), genetic algorithms (GA), particle swarm optimization (PSO), Covariance Matrix Adaptation Evolution Strategy (CMA-ES) and other policy learning methods. This work has already conducted to the development of an efficient set of humanoid low-level skills. Section 6 presents briefly our new learning framework. We have also developed new walking and running models for humanoid robots that emphasize speed, stability and flexibility [31].

## 2. Research Directions

New research directions include research on developing our current layered architectures for agent controlling to be optimized and more efficient. Thus our research will be focused on improving both lower layers and higher layers. The lower layers will be responsible for the basic control of the humanoid such as stability, while the higher layers take decisions at a strategic level.

In our lower level control architecture, we will develop new kick skill based on our previous kick skill; also we will improve our developed running skill to be used as a primary locomotion for our robots. The robustness of the walking and running skill in face of the external forces will also be improved.

In the upper level control architecture, directions of research in FC Portugal include developing a model for a strategy for a humanoid game and the integration of humanoids coming from different teams in an inter-team framework to allow the formation of a team with different humanoids, and developing a new opponent modeling approach to model the opponent basic behaviors performance, its positioning, etc. These are factors that must be taken into account when selecting a given strategy for a game.

One of the improvements that have been achieved during the previous years was implementing a new learning framework that allows us to more efficiently optimize our agent behaviours using state-of-the-art deep reinforcement learning algorithms, such as Proximal Policy Optimization (PPO) and Asynchronous Advantage Actor Centralized-Critic with Communication (A3C3). These techniques have been combined with physics based models and optimum control to derive very efficient skills.

Also heterogeneity will be important because in the future it is expected that not all humanoids will be identical, having humanoids with different capabilities introduces new problems of task assignment that will have to be dealt with in humanoid teams. We have already tested the use of heterogeneous humanoids in the 2014 3D Simulation competition. We optimize behavior specifications for each heterogeneous humanoid robot.

## 3. Agent Architecture

The FC Portugal Agent 3D [7] is divided in several packages: each one with a specific purpose. Figure 1 shows the general structure of the humanoid agent.

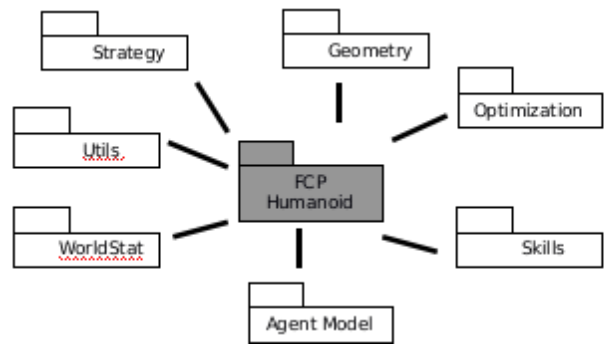

**Fig. 1:** FCP Humanoid Agent Architecture

- **World State**: Contains classes to keep track of the environment information. These include the objects presented in the field (fixed objects as is the case of flags and goals and mobile objects as is the case of the players and the ball), the game state, (e.g. time, play mode) and game conditions (e.g. field length, goals length);

- **Agent Model**: Contains a set of classes responsible for the agent model information. This includes the body structure (body objects such as joints, body parts and perceptors), the kinematics interface, the joint low-level control and trajectory planning modules;
- **Geometry**: Contains useful classes to define geometry entities as is the case of points, lines, vectors, circles, rectangles, polygons and other mathematical functions;
- **Optimization**: Contains a set of classes used for the optimization process. These classes are a set of evaluators that know how each behavior should be optimized;
- **Skills**: This package is associated with the reactive skills and talent skills of the agent. Reactive skills include the base behaviors as is the case of walk in different directions, turn, get up, kick the ball and catch the ball. Talent skills are some powerful think capabilities of the agent, which include movement prediction of mobile objects in the field and obstacle avoider;
- **Utils**: This package is related with useful classes that allow the agent to work. This includes classes for allowing the communication between the agent and the server, communication between agents, parsers and debuggers.
- **Strategy**: Contains all the high-level functions of the agent. The package is very similar to the team strategy packages used for other RoboCup leagues.

## 4. Low-Level Skills

In this section we briefly describe our approaches to develop soccer low-level skills, such as walking, running, and kicking. Nowadays, in order to compete well in RoboCup soccer humanoid leagues, the robots should be able to perform their low level skill fast, in an omni-directional manner, and also robust against the external perturbation and noises. In order to improve our low-level skills, we model the dynamics of them by using simple physical system. Then we try to apply the optimization techniques, in order to tune parameters of those models of that skill.

### 4.1 Modelling and Controlling the Dynamics of the Low-Level Skills

Many popular approaches used for controlling the balance of bipedal locomotion are based on the Zero Momentum Point (ZMP) stability indicator and inverted pendulum model. ZMP cannot generate reference walking trajectories directly but it can indicate whether generated walking trajectories will keep the balance of a robot or not. Kajita et al. assumed that biped walking is a problem of balancing a cart-table model [8], since in the single supported phase, human walking can be represented as the Cart-table model. Biped walking can be modeled through the movement of ZMP and CoM. The robot is in balance when the position of the ZMP is inside the support polygon. When the ZMP reaches the edge of this polygon, the robot loses its balance. Cart-table model has some assumptions and simplifications in its model. One the major drawback of the cart-table model is its consideration the height of the robot fixed during it movement, which is not true for many soccer low-level skills such as running or kicking, Therefore we used the inverted pendulum model which does not have this issue. Figure 2 shows how robot dynamics is modeled by an inverted pendulum and its schematic view.

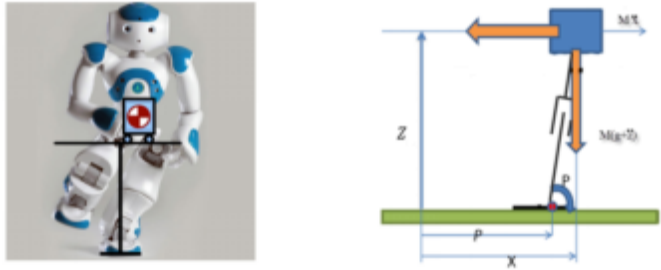

**Fig. 2**. Schematic view of inverted pendulum model on X direction and frontal view of the NAO robot

Two sets of inverted pendulum are used to model 3D walking. One is for movements in frontal plane; another is for movements in coronal plane. The position of Center of Mass (CoM) $M$ is $x$ and $z$ defined in the coordinate system $O$. Gravity $g$ and cart acceleration create a moment $T_p$ around the center of pressure (CoP) point $P_x$. The Equation (1) provides the moment or torque around $P$.

$$T_p = M(g + \ddot{z})(x - P_x) - M\ddot{x}z \quad (1)$$

We know from [9] that when the robot is dynamically balanced, ZMP and CoP are identical, therefore the amount of moment in the CoP point must be zero, $T_p=0$. By assuming the left hand side of equation (1) to be zero, equation (2) provides the position of the ZMP. Another cart-table must be used in y direction. Using the same assumption and reasoning equation (3) can be obtained. Here, y denotes the movement in y.

$$P_x = x - \frac{z}{g + \ddot{z}}\ddot{x} \quad (2)$$

$$P_y = y - \frac{z}{g + \ddot{z}}\ddot{y} \quad (3)$$

In order to apply inverted model in a biped walking problem, first the position of the support foot during the performing the low-level skill must be planned and defined, then based on the constraint of ZMP position and support polygon, the ZMP trajectory can be designed. In the next step, the position of the CoM must be calculated using differential equations (2) (3). One of the main issue of using the inverted pendulum, is how to solve these differential equations or how to generated trajectory. We presented an approach for the solution of the cart- table model analytically in [13]. However the solution of the inverted pendulum model cannot be derived analytically, instead recently we have presented a numerical approach to solve inverted pendulum model and to calculate the CoM trajectory. This approach is explained in detains in [14]. Finally, inverse kinematics is used to find the angular trajectories of each joint based on the planned position of the foot and calculated CoM trajectory. We used our two different inverse kinematic approaches, which were applied on the NAO humanoid soccer robot can be found in [10] [11].

**4.2 Omni-Directional Biped Locomotion**

This section briefly presented the design of the locomotion controllers to enable the robot with an omni-directional walking. We use this design to implement walking approached by using both inverted pendulum and cart–table model. The extended details of our approach can be found in [13].
Developing an omni-directional biped locomotion is a complex task made of several components. In order to get a functional omnidirectional walk, it is necessary to decompose it in several modules and address each module independently. Each modules is explains in the following.

- **ZMP Trajectory Generator -** In this module the ZMP generated using the desired velocities. This computation takes into account only the linear component of the walk, which means to walk in any direction always looking to the same direction, like diagonal walk.
- **Foot Planner -** One of the drawbacks of the linear inverted pendulum model is the need for a constant height. We can improve this by adjusting the CoM height using the length of the leg, support foot position and the ground projection of the CoM. In [12] a detailed explanation is given.
- **CoM Trajectory generator –** This module is responsible to Generate the CoM by using the dynamics equations of inverted pendulum model [14], or cart-table model [12] [13].
- **Swing Trajectory Generator -** This module is responsible to generate a trajectory for the swing foot. It uses the cycloid parametric equation to generate the desired trajectory.
- **Feet Frame Computation -** After computing the support foot (ZMP) position, CoM position and swing trajectory feet position has to be computed taking into account if it is in double support phase or single (left or right) support phase. This module is responsible for computing the position and orientation of both feet relative to the CoM frame.
- **Active Balance -** This module is where the balance of the humanoid during locomotion is controlled in order to maintain it stable. The inverted pendulum model has some simplifications in biped walking dynamics modeling; in addition, there is inherent noise in leg's actuators. Therefore, keeping walk balance generated by inverted pendulum model cannot be guaranteed. In order to reduce the risk of falling during walking, an active balance technique is applied. The detailed explanation of this module can be found in [12].

The Active balance module tries to keep an upright trunk position by decreasing variation of trunk angles. One PD controller is designed to control the trunk angle to be the desired trunk pitch angle. An inertial measurement unit which is included in the robot body gives the trunk inclination angle. When the trunk angle is not the desired pitch angle, instead of considering a coordinate frame attached to the trunk of the biped robot, position and orientation of the feet are calculated with respect to a coordinate frame, which is attached to the CoM position and the *Z* axes always has the predefined pitch angle to the ground plane. For example if the trunk pitch offset is assumed to be zero, the *Z* axes keeps always perpendicular to the ground plane.
The PD controller calculates the rotation angle based on the difference between the current trunk inclination and the desired trunk pitch angle. The calculated rotation angle is a portion of this difference and the coordinate frame rotates with the calculated rotation angle. By using this transformation, the controller tries to keep the Z axis of the coordinate frame in a desired angle to the ground plane. The foot position is calculated by using the rotated coordinated frame, the feet orientation also tries to be kept parallel to the ground.

The Transformation formulation is presented in equation (4).

$$Foot = T^{CoM}_{Foot}(pitchAng, rollAng) \times \overline{Foot} \quad (4)$$

The *pitchAng* and *rollAng* are assumed to be the angles calculated by the PID controller around *y* and *x* axis respectively. Figure 3 shows the architecture of the active balance unit when the trunk pitch offset is assumed to be zero.

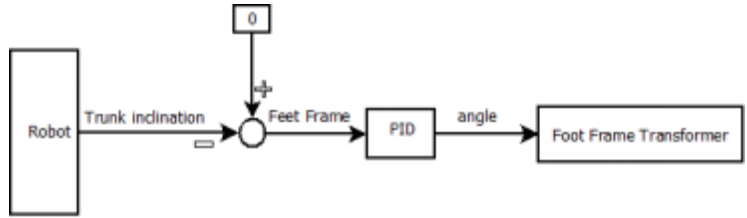

**Fig. 3:** Active Balance controller

We have also begun work in a walking engine for humanoid robots which uses a hybrid ZMP-GPC framework to allow for fast omnidirectional walk, while still being robust to perturbations [30]. Despite promising initial results, the speed of this new walking engine is still beneath our previous work.

**4.3 Model-free running**

We developed a way of leveraging reinforcement learning and the information provided by the simulator for official RoboCup matches to learn human-like running skills. To do this, our algorithm uses a mix of raw, computed and internally generated data, as shown in Figure 4.

| Parameter | Data size ($\times$32b) | Acquisition Method |
|---|---|---|
| counter | 1 | generated |
| z-coordinate | 1 | computed |
| orientation* | 1 | computed |
| gyroscope* | 3 | raw |
| accelerometer* | 3 | raw |
| feet force* | 12 | raw |
| joint's position* | 20 | raw |
| *differentiation | 39 | computed |

*differentiation is performed on parameters followed by an asterisk

**Fig. 4:** Reinforcement learning state space representation

The counter is generated by the agent. It starts from zero, when the robot is stopped, and it is incremented at every 3 time steps (frequency of visual information). This counter works like a timer for the optimization algorithm. The z-coordinate provides useful information when the feet data is unavailable and the orientation is essential to keep the robot aligned with the objective. The former is easily determined from the visual references provided by the simulator. The latter is computed with a linear predictor function that was modeled using linear regression on some of the other state space variables. The best single features and features crosses were initially filtered based on their correlation with the ground-truth, which was obtained directly from SimSpark. The features were then refined and used to optimize the regression model. In addition to the robot's sensors and joints, a first order derivative of all above mentioned parameters is also computed.

Two main skills were developed – run and sprint. The former allows the robot to turn to any direction while running. It is composed of a main action and two subtasks which allow the robot to stop or progressively shift to walking. The latter skill is more focused on speed and less on turning, and, due to its flexible nature, it can end with a ball kick, in addition to stopping or shifting to walking. Both behaviors control all the NAO's joints except for the head. The resulting models dictate the joints angles, consequently controlling their speed through a proportional controller. They were learned using Proximal Policy Optimization (PPO) – a model-free reinforcement learning algorithm. The optimization was performed for 200M time steps using the SimSpark simulator.

Figure 5 shows relevant statistics for the most successful robot types. The displayed values for the main skills were averaged over 1000 episodes of 10 seconds each. A video demonstration is available online at https://youtu.be/lkSVad tjOY.

| Skill | Avg. & Max. linear speed along $x$ | Max. rot. speed | Subtask | Duration |
|---|---|---|---|---|
| Sprint | 2.48m/s & 2.62m/s$ | $10°/s$ | Walk Transition | $0.9s$ |
| | | | Stop | $[1, 1.8]s$ |
| Run | $1.41m/s$ & $1.52m/s$ | $160°/s$ | Walk Transition | $0.9s$ |
| | | | Stop | $[1, 1.6]s$ |
| | | | Kick | N.A. |

**Fig. 5:** Sprinting, running and stopping results

### 4.4 Humanoid Kick with Controlled Distance

We investigate the learning of a flexible humanoid robot kick controller, i.e., the controller should be applicable for multiple contexts, such as different kick distances, initial robot position with respect to the ball or both. Current approaches typically tune or optimise the parameters of the biped kick controller for a single context, such as a kick with longest distance or a kick with a specific distance. Hence our research question is "how can we obtain a flexible kick controller that controls the robot (near) optimally for a continuous range of kick distances?". The goal is to find a parametric function that given a desired kick distance, outputs the (near) optimal controller parameters. We achieve the desired flexibility of the controller by applying a contextual policy search method. With such a contextual policy search algorithm, we can generalize the robot kick controller for different distances, where the desired distance is described by a real-valued vector [29].

Figure 8 shows an example of an initial and final stance for the kick behavior. Our movement pipeline is composed of two main parts: a kick controller, which receives parameters θ and converts them into joint commands for the robot's servos; and a policy function, which maps a given context s for a specific kick distance into the corresponding parameter vector θ. The pipeline for the kick task, whose context is the kick distance s with a straight kick direction with respect to the torso, is shown in Figure 9.

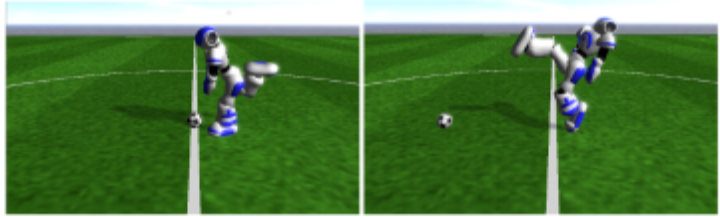

**Fig 8.** The initial (left) and final (right) positions of an exemplary kick movement.

In order to learn the policy function (s) we use a contextual policy search algorithm called CREPS-CMA [28].

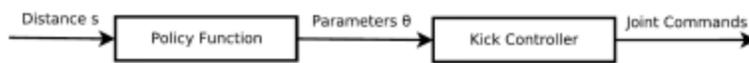

**Fig 9.** The pipeline of our contextual kick movement.

We use CREPS-CMA to train the 3D simulated NAO robot by optimising the kick controller. The desired kick distance *s* varies from 2:5m to 12:5m. For the non-linear policy, we choose K = 15 normalized RBFs and $\sigma^2$ is set to 0.5. Both K and the $\sigma^2$ parameters were chosen by trial and error to maximize the results accuracy.

We achieved an average error of 0.34±0.11*m* using the non-linear policy. Figure 10 shows the learned non-linear policies for generalizing the 25 parameter kick controller for different kick distances.

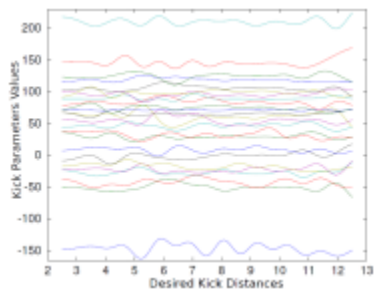

**Fig 10.** The learned non-linear policy for kick distances of 2.5 to 12.5 meters. The y-axis represents the controller parameter values for a given desired kick distance, and the x-axis represents the desired kick distance.

## 5. High-Level Decisions and Coordination

Flexible Tactics has always been one of the major assets of FC Portugal teams. FC Portugal 3D is capable of using several different formations and for each formation players may be instantiated with different player types. The management of formations and player types is based on SBSP – Situation Based Strategic Positioning algorithm [1, 4]. Player's abandon their strategic positioning when they enter a critical behavior: Ball Possession or Ball Recovery. This enables the team to move in a quite smooth manner, keeping the field completely covered.

The high-level decision uses the infrastructure presented in Section 3. Several new types of actions are currently being considered taking in consideration the new opportunities opened by the 3D environment of the new simulator. We also have adapted our previously researched methodologies to the new 3D environment:

- Strategy for a Competition with a Team with Opposite Goals [1, 4, 5, 21];
- Concepts of Tactics, Formations and Player Types [1, 3, 4, 21];
- Distinction between Active and Strategic Situations [1, 4];
- Situation Based Strategic Positioning (SBSP) [1, 4, 5];
- Dynamic Positioning and Role Exchange (DPRE) [1, 4, 5];
- Visual Debugging and Analysis Tools [1, 3, 22];
- Optimization based Low-Level Skills [1, 3, 26, 27].
- Standard Language to Coach a (Robo)Soccer Team[2,3];
- Intelligent Communication using a Communicated World State [1, 3, 5];
- Flexible Set-plays for coordinating robosoccer teams [23];
- Generic reinforcement learning framework;
- Model-free running skills [32].

In previous years, our research was mostly concerned in developing optimization based low level skills for the humanoid agent and robust mid-level skills. The high-level layers of the team for 2020 will be adapted to be used in the humanoid simulator (these methodologies have already been adapted to our Simulation 2D, Simulation 3D with spheres model, small-size, middle-size [24] and rescue teams [25]).

**6. Learning Framework**

For developing a learning framework, it is better to run the simulation as fast as the CPU can. By using *SyncMode*, the simspark simulator only waits for the agent commands and a synchronize message which signals the end of the agent cycle. After it receives all the agents Sync message, the server processes all the commands and proceeds to the next cycle. In addition to simulation speed time improvement, it can also be used to detect strange cycle times from the agents. We have developed our agent to use SyncMode to improve the speed of the optimization process.

The learning framework is based on OpenAI Gym, as it provides a very simple to use interface and is generic enough to allow any algorithm or computational library to be used. Most of our optimization tasks have been using the PPO algorithm and Tensorflow for the neural networks optimization.

The framework also completely hides our agent code, which allows us to share it as a library to students and other people not directly related to FCPortugal3D, thus increasing student contact and promoting the Robocup scene.

**6. Conclusions**

Robust low-level skills have been developed for the NAO humanoid model, the results of low-level skills have already tested and validated on the real NAO robot, since it is based on the physical modeling of the dynamics of biped locomotion it is very robust and with minimal adaptation was used on the NAO robot. Using optimization and learning techniques, enabling us to continue the research in strategical reasoning and coordination methodologies that should be the focus of the simulation leagues inside RoboCup. Also the extended flexibility of omnidirectional kicks and walks will enable a more cooperative game style.

Future work will be concerned in extending the optimization methodology for skills sequences and on developing coordination methodologies enabling teams of humanoid robots to play robosoccer games in a robust and flexible manner. This includes new deep learning techniques for both low and high level behaviors.

Almost all of our research on high-level flexible coordination methodologies is directly applicable to the 3D league and the increase in the number of elements of the each team is very welcome, enabling coordination methodologies to be useful in this league. FC Portugal started its participation in the SPL - Standard Platform League in 2011. The SPL code of the team is entirely made from scratch based on the Simulation 3D code. Thus, future work will be on bridging the gap between simulation and robotics by developing a more realistic NAO model in Simspark enabling better portability of the simulated code to the real robot.

**Acknowledgements**

This work was partially supported by the Portuguese National Foundation for Science and Technology: SFRH/BD/66597/2009 and SFRH/BD/81155/2011. This research was partially supported by IEETA (UID/CEC/00127/2019) and LIACC (PEst-UID/CEC/00027/2019).